\documentclass[conference]{IEEEtran}
\IEEEoverridecommandlockouts
\usepackage{cite}
\usepackage{amsmath,amssymb,amsfonts}
\usepackage{algorithmic}
\usepackage{graphicx}
\usepackage{textcomp}
\usepackage{xcolor}
\usepackage{float}
\usepackage{multirow}
\usepackage{adjustbox}
\usepackage[normalem]{ulem}
\useunder{\uline}{\ul}{}
\def\BibTeX{{\rm B\kern-.05em{\sc i\kern-.025em b}\kern-.08em
    T\kern-.1667em\lower.7ex\hbox{E}\kern-.125emX}}
\begin{document}

\title{CVSS-BERT: Explainable Natural Language Processing to Determine the Severity of a Computer Security Vulnerability from its Description\\
}

\author{\IEEEauthorblockN{Mustafizur R. Shahid}
\IEEEauthorblockA{\textit{SAMOVAR, Télécom SudParis} \\
Institut Polytechnique de Paris\\
France \\
mustafizur.shahid@telecom-sudparis.eu}
\and
\IEEEauthorblockN{Hervé Debar}
\IEEEauthorblockA{\textit{SAMOVAR, Télécom SudParis} \\
Institut Polytechnique de Paris\\
France \\
herve.debar@telecom-sudparis.eu}
}

%\author{\IEEEauthorblockN{Mustafizur R. Shahid}
%\IEEEauthorblockA{\textit{Institut Polytechnique de Paris} \\
%France \\
%mustafizur.shahid@telecom-sudparis.eu}
%\and
%\IEEEauthorblockN{Hervé Debar}
%\IEEEauthorblockA{\textit{Institut Polytechnique de Paris} \\
%France \\
%herve.debar@telecom-sudparis.eu}
%}

%\and

%}

\maketitle

\begin{abstract}
When a new computer security vulnerability is publicly disclosed, only a textual description of it is available. Cybersecurity experts later provide an analysis of the severity of the vulnerability using the Common Vulnerability Scoring System (CVSS). Specifically, the different characteristics of the vulnerability are summarized into a vector (consisting of a set of metrics), from which a severity score is computed. However, because of the high number of vulnerabilities disclosed everyday this process requires lot of manpower, and several days may pass before a vulnerability is analyzed. We propose to leverage recent advances in the field of Natural Language Processing (NLP) to determine the CVSS vector and the associated severity score of a vulnerability from its textual description in an explainable manner. To this purpose, we trained multiple BERT classifiers, one for each metric composing the CVSS vector. Experimental results show that our trained classifiers are able to determine the value of the metrics of the CVSS vector with high accuracy. The severity score computed from the predicted CVSS vector is also very close to the real severity score attributed by a human expert. For explainability purpose, gradient-based input saliency method was used to determine the most relevant input words for a given prediction made by our classifiers. Often, the top relevant words include terms in agreement with the rationales of a human cybersecurity expert, making the explanation comprehensible for end-users.
\end{abstract}

%\begin{IEEEkeywords}
%NLP, explainability, cybersecurity
%\end{IEEEkeywords}

\section{Introduction}

A computer security vulnerability can be a bug, a flaw or a weakness that can be exploited by a malicious actor to cause a failure of the confidentiality, the availability or the integrity of the system.  A zero-day vulnerability is a computer security flaw known to a limited number of parties (the software vendor or cybercriminals) but unknown to the general public. When the existence of a vulnerability is disclosed to the public, software patches might not be available yet. In fact, it is not uncommon to have a significant delay between the disclosure of a vulnerability and the moment a patch or a security fix is made available by the vendor. Even when a security patch is available at disclosure, it might not have been deployed to all the affected systems. Early vulnerability scoring might also be approximative.

Thousands of vulnerabilities are disclosed every year. Most organizations do not have the resources (time, manpower, etc.) to address all the disclosed vulnerabilities that affect their systems immediately. Instead, they must prioritize their efforts. Moreover, patching complex enterprise systems might cause significant downtime and unwanted side effects. Therefore, it is necessary for system administrators to determine which vulnerabilities should be addressed first. Knowing the severity of a vulnerability might help them to prioritize their efforts and allocate resources accordingly.

New vulnerabilities are disclosed through the Common Vulnerabilities and Exposures (CVE)~\cite{CVE_MITRE} system. CVE is a list of records of publicly disclosed computer security vulnerabilities, operated and maintained by MITRE. An entry in the CVE list contains an identification number (CVE ID), a description of the vulnerability, and at least one public reference (links to vulnerability reports, advisories, etc.). Next, the NIST National Vulnerability Database (NVD)~\cite{NVD_NIST} builds upon the information provided by CVE records to provide enhanced information for each record such as fix information, severity scores, and impact ratings.  Those additional knowledge about vulnerabilities found in NVD are provided by human security experts.  An example of additional information provided by NVD is an analysis of the severity of a vulnerability in the form of a vector and a score using the Common Vulnerability Scoring System (CVSS)~\cite{CVSSv3.1}. The CVSS provides a way to summarize the principal characteristics of a vulnerability through a vector that contains a set of metrics on how easy it is to exploit the vulnerability (exploitability metrics) and the impact of a successful exploit (impact metrics). A numerical score is computed from the CVSS vector to assess the severity of a vulnerability relative to other vulnerabilities. The process of assessing a newly disclosed vulnerability, and attributing a CVSS vector to it, requires expert knowledge. Because of the high number of vulnerabilities disclosed everyday, this process might require a lot of time and manpower. In some cases, it can take days before a newly disclosed vulnerability is analyzed by NVD security experts and attributed a CVSS vector.

Our contribution leverages recent advances in the field of Natural Language Processing (NLP) to determine the CVSS base vector and the associated severity score of a vulnerability from its textual description provided by CVE, in an automated and explainable way. We use BERT (Bidirectional Encoder Representations from Transformers)~\cite{devlin2018bert}, a transformer-based language representation model. Multiple BERT classifiers are trained, each to determine the value of a specific metric composing the CVSS base vector (AC, AV, PR, UI, S, C, I, A see Section~\ref{CVSS} for detailed information about each metric). The severity score of the vulnerability is then computed from the predicted CVSS vector. Explainability is an important requirement for our system. It allows the end-users to understand the decision of our model and justify the predicted CVSS vectors and severity scores. It is also useful to debug the model and for knowledge discovery. Hence, we propose to use gradient-based input saliency method to find out which words in the textual description of a vulnerability were the most relevant for a given prediction made by our model. We also use this method to discover which words are most often associated by our trained models with specific values of the metrics composing the CVSS vector. For example, for the classifier trained to predict the Confidentiality Impact metric of the CVSS vector, we determine which words and bigrams in vulnerability descriptions most often lead it to predict HIGH, LOW or NONE.

\section{Related Work}

A limited number of works on vulnerability severity prediction exists. 

C. Elbaz et al.~\cite{elbaz2020fighting} propose to predict the metrics of the CVSS base vector as well as the associated severity score from the description of a vulnerability. The description of a vulnerability is transformed into a bag of words. A bag of words is a vector with each dimension corresponding to the  number of occurrences of a given word (0 indicating absence). Irrelevant words are removed to reduce the dimension  of the vulnerability vector.  Linear regression models are trained to predict a score for each metric of the CVSS vector.  The value of each metric of the CVSS vector are then inferred from the predicted numerical scores. The use of simple linear regression models has the advantage of maintaining some level of explainability, as the weight of each word in the prediction can help to determine the most relevant words. However, linear regression assumes linear relationship between the input and the output. Therefore, it fails to properly model the complexity of natural language, limiting the performance of the model. Moreover, bag of words representation ignores context and discards words ordering, resulting in a poor representation of text data.

A. Khazaei et al.~\cite{khazaei2016automatic} propose to predict discretized approximate CVSS severity sores from vulnerability descriptions. The input data is created as follows: stop words are removed from the descriptions, the remaining words are stemmed, the TF-IDF (Term Frequency–Inverse Document Frequency) value of each word is calculated. The output of the model is  a discretized  CVSS score: the continuous CVSS score interval range [0, 10] is  divided into 10 equal sub intervals, each corresponding to a different class. Hence, the problem is a 10-class classification problem. Three different models are trained and tested, SVM and Random Forest with a dimensionality reduction step, and a fuzzy system. The presented approach does not reconstruct the full CVSS vector and attempts only to provide an approximate severity score. The authors also do not provide any way to explain the results predicted by the model.

Z. Han et al.~\cite{han2017learning} propose to predict qualitative CVSS severity ratings (Low, Medium, High, Critical) from vulnerability description. First, to represent words in a vector space, word embeddings are trained using a continuous skip-gram models. Word embeddings attempt to encode the meaning of words and are a type of word representation that allows words with similar meaning to be close in the vector space. A vulnerability description is transformed into a set of vectors consisting of the concatenation of word embeddings of words present in the description.  The obtained representation is fed to a Convolutional Neural Network (CNN) to determine the severity ratings of the vulnerability. The work does not aim at reconstructing the full CVSS vector. It only attempts to provide a categorical severity rating (and not a precise numerical severity score). The proposed model is a black-box and the authors do not provide any way to explain the predictions.

Other related works worth mentioning include~\cite{zong2019analyzing,zou2019autocvss,tavabi2018darkembed,russo2019summarizing,jacobs2019exploit}. S. Zong et al. ~\cite{zong2019analyzing} analyze the perceived cybersecurity threat reported on social media in an attempt to predict the real severity of a vulnerability. In~\cite{zou2019autocvss} the severity of a vulnerability is determined based on attack process (the corresponding proof of concept exploit and vulnerable software). N. Tavabi et al.~\cite{tavabi2018darkembed} analyze darkweb/deepweb discussions to predict whether vulnerabilities will be exploited. Similarly, ~\cite{jacobs2019exploit} describes a method to determine the probability that a vulnerability will be exploited in the wild within the first twelve months after its public disclosure. ~\cite{russo2019summarizing} presents an approach able to automatically generate summaries of daily posted vulnerabilities and categorize them according to a taxonomy modeled for the industry. %Example of categories include, affected software, versions, type of attackers, origin of the vulnerability and its consequences. 

\section{Attention Based NLP Models}

We first present the Transformer architecture and attention mechanisms. Then, we describe BERT and how it can be used for classification task.

\subsection{Transformers}

\begin{figure}[]
\begin{center}
\includegraphics[scale=0.39]{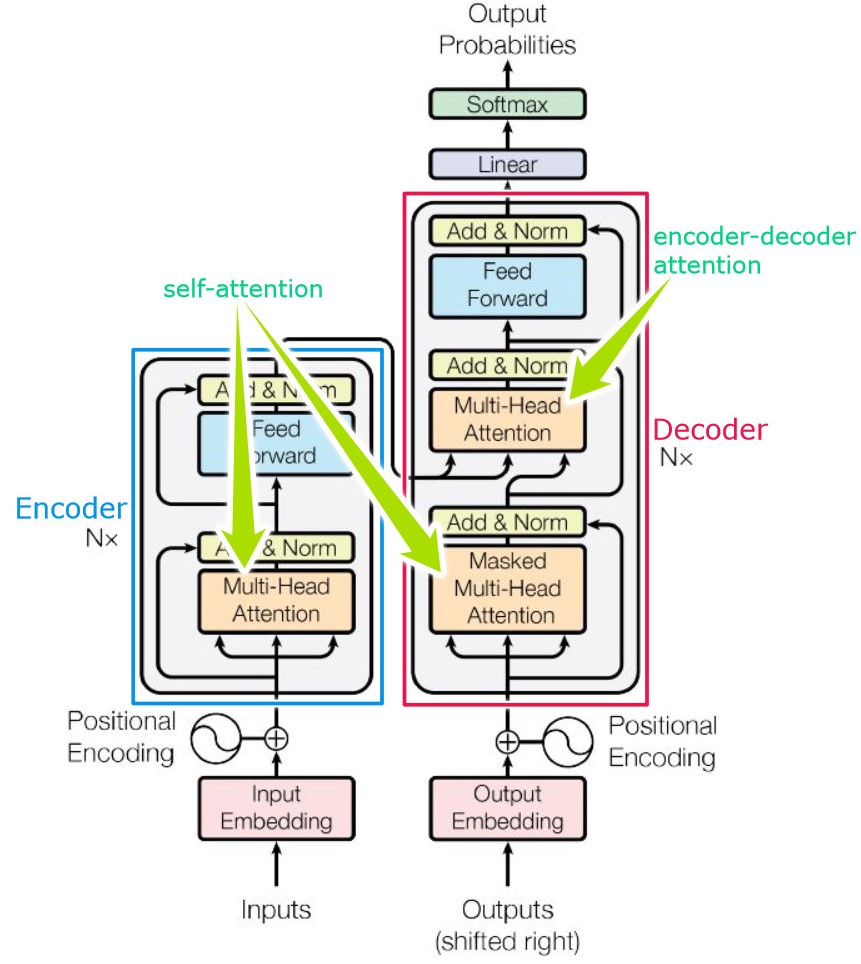}
\end{center}
\caption{The Transformer - model architecture}
\label{Transformer}
\end{figure}

Vaswani et al.~\cite{vaswani2017attention} proposed the Transformer, which significantly improved the performance of Neural Machine Translation (NMT) applications, and is faster to train and easier to parallelize. As illustrated in Figure~\ref{Transformer}, to process an input sentence, represented by a sequence of words, Transformers do not use any recurrent or convolutional layers but rely on attention mechanisms. As, the Transformer was designed for NMT, it consists of an \textit{encoder} and a \textit{decoder}. Actually, the encoder is a stack of N=6 encoders, and the decoder is also a stack of N=6 decoders. Lets consider an NMT application that translates English sentences (inputs) to their French equivalents (outputs). The stack of encoders encode each English input words into an internal representation. The stack of decoders outputs the translated French sentence word by word. To output the next word, it takes as input the encoded representation of the original English sentence, along with the words translated so far.

The Transformer architecture is composed of modules like fully connected feed-forward networks, residual connections or layer normalizations that are commonly found in other neural network architectures. The main novelty introduced are the different types of \textit{attention} layers described hereafter.

An encoder consists of a \textit{multi-head self-attention layer}. The role of a self-attention layer is to quantify the interdependence within the words of an input sentence. It encodes the relationship between each word of a sentence, with every other words of the same sentence. For example, in the sentence “The animal didn't cross the street because it was too tired”, self-attention attention allows the model to associate the word “it” with the word “animal”. Put another way, the word “it” will pay more attention to the word “animal” than to any other words of the sentence.

A decoder consists of a \textit{masked multi-head self-attention layer} and an \textit{multi-head encoder-decoder attention layer}.  A masked self-attention layer does the same thing as the self-attention layer used in an encoder, except that each word is allowed to only attend the words before it. The role of an encoder-decoder attention layer is to quantify the interdependence between the words of an input sentence and the words of an output sentence. It encodes each output word’s relationship with every words of the input sentence. For example, when translating the English sentence “how are you?” into its French equivalent “Comment allez-vous?” , the encode-decoder attention allows the model to associate the French word “Comment” to the English equivalent “How”. Put another way, when translating the word “Comment”, the decoder will pay more attention to the word “How” than to any other words in the original English sentence.

\subsection{BERT} \label{BERT}

\begin{figure}[]
\begin{center}
\includegraphics[scale=0.45]{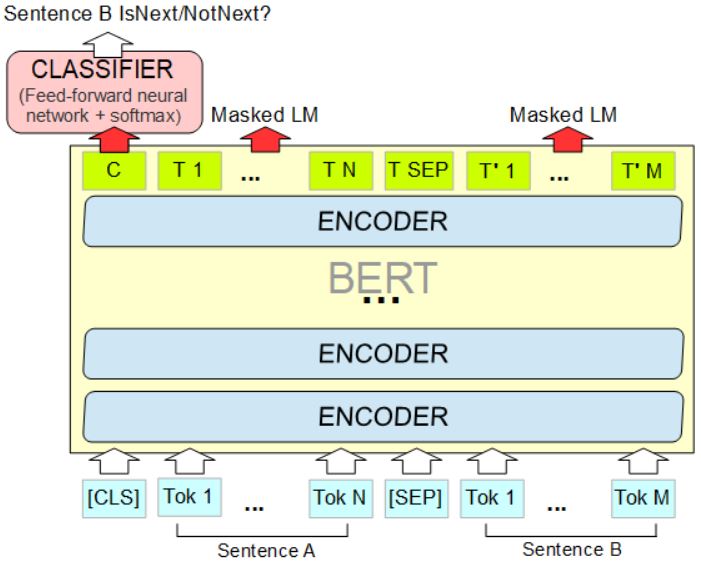}
\end{center}
\caption{BERT pretraining}
\label{BERT_pretrain}
\end{figure}

\begin{figure}[]
\begin{center}
\includegraphics[scale=0.45]{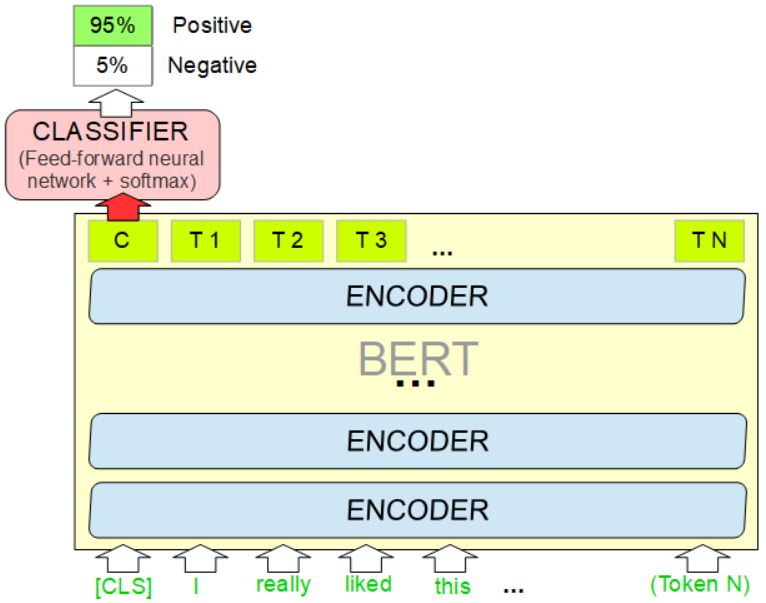}
\end{center}
\caption{BERT for classification task}
\label{BERT_class}
\end{figure}

% \usepackage{multirow}
% \usepackage[normalem]{ulem}
\begin{table*}[h!]
\centering
\caption{CVSS Base metrics}
\label{tab:cvss_metrics}
\begin{tabular}{|c|c|l|l|}
\hline
                                                                                  &                                                                    & \multicolumn{1}{c|}{Description}                                                                                                                                                                                                                                                                     & \multicolumn{1}{c|}{Possible Values}                                                          \\ \hline
\multirow{5}{*}{\begin{tabular}[c]{@{}c@{}}Exploitability\\ Metrics\end{tabular}} & \begin{tabular}[c]{@{}c@{}}Attack Vector\\ (AV)\end{tabular}       & \begin{tabular}[c]{@{}l@{}}Reflects the context by which vulnerability exploitation is possible. This\\ metric value will be larger the more remote an attacker can be in order\\ to exploit the vulnerable component.\end{tabular}                                                                  & \begin{tabular}[c]{@{}l@{}}Network (N)\\ Adjacent (A)\\ Local (L)\\ Physical (P)\end{tabular} \\ \cline{2-4} 
                                                                                  & \begin{tabular}[c]{@{}c@{}}Attack\\ Complexity\\ (AC)\end{tabular} & \begin{tabular}[c]{@{}l@{}}Describes the conditions beyond the attacker’s control that must exist in\\ order to exploit the vulnerability. Such conditions may require the collection\\ of more information about the target, or computational exceptions.\end{tabular}                              & \begin{tabular}[c]{@{}l@{}}Low (L)\\ High (H)\end{tabular}                                    \\ \cline{2-4} 
                                                                                  & \begin{tabular}[c]{@{}c@{}}Privileges\\ Required (PR)\end{tabular} & \begin{tabular}[c]{@{}l@{}}Describes the level of privileges an attacker must possess before\\ successfully exploiting the vulnerability.\end{tabular}                                                                                                                                               & \begin{tabular}[c]{@{}l@{}}None (N)\\ Low (L)\\ High (H)\end{tabular}                         \\ \cline{2-4} 
                                                                                  & \begin{tabular}[c]{@{}c@{}}User Interaction\\ (UI)\end{tabular}    & \begin{tabular}[c]{@{}l@{}}Captures the requirement for a human user, other than the attacker, to\\ participate in the successful compromise of the vulnerable component.\end{tabular}                                                                                                               & \begin{tabular}[c]{@{}l@{}}None (N)\\ Required (R)\end{tabular}                               \\ \cline{2-4} 
                                                                                  & Scope (S)                                                          & \begin{tabular}[c]{@{}l@{}}Captures whether a vulnerability in one vulnerable component impacts\\ resources in components beyond its security scope.\end{tabular}                                                                                                                                    & \begin{tabular}[c]{@{}l@{}}Changed (C)\\ Unchanged (U)\end{tabular}                           \\ \hline
\multirow{3}{*}{\begin{tabular}[c]{@{}c@{}}Impact\\ Metrics\end{tabular}}         & Confidentiality (C)                                                & \begin{tabular}[c]{@{}l@{}}measures the impact to confidentiality of a successfully exploited\\ vulnerability. Confidentiality refers to limiting information access and\\ disclosure to only authorized users, as well as preventing access by, or\\ disclosure to, unauthorized ones.\end{tabular} & \begin{tabular}[c]{@{}l@{}}High (H)\\ Low (L)\\ None (N)\end{tabular}                         \\ \cline{2-4} 
                                                                                  & Integrity (I)                                                      & \begin{tabular}[c]{@{}l@{}}Measures the impact to integrity of a successfully exploited vulnerability.\\ Integrity refers to the trustworthiness and veracity of information.\end{tabular}                                                                                                           & \begin{tabular}[c]{@{}l@{}}High (H)\\ Low (L)\\ None (N)\end{tabular}                         \\ \cline{2-4} 
                                                                                  & Availability (A)                                                   & \begin{tabular}[c]{@{}l@{}}measures the impact to availability of a successfully exploited vulnerability.\\ This metric refers to the loss of availability of the impacted component\\ itself, such as a networked service (e.g., web, database, email).\end{tabular}                                & \begin{tabular}[c]{@{}l@{}}High (H)\\ Low (L)\\ None (N)\end{tabular}                         \\ \hline
\end{tabular}
\end{table*}

In~\cite{devlin2018bert}, J. Devlin et al. proposed BERT (Bidirectional Encoder Representations from Transformers) designed to learn bidirectional representations from unlabeled text by jointly conditioning on both left and right context (instead of reading text sequentially, from left-to-right or from right-to-left). BERT has been pretrained on two tasks: masked language model (MLM) and next sentence prediction (NSP).
As shown in Figure~\ref{BERT_pretrain}, BERT is basically a Transformer encoder stack. Sentences are tokenized before being fed to the model. Tokens can represent words or subwords. For example, some long words or words that are uncommon might be represented using multiple tokens. The first input token is a special token [CLS] that will be used for the NSP task. The model outputs vectors, one for each input token. For the MLM task, some input tokens are masked and the model is trained to predict them. This is accomplished by feeding the output vector corresponding to a masked token to a fully connected feed-forward neural network that output a softmax over the vocabulary. For the NSP task the model is fed with two sentences A and B and it has to predict whether or not sentence B follows sentence A. This is done by feeding the output vector corresponding to the special [CLS] token to a fully connected feed-forward neural network that performs binary classification. BERT was pretrained using text extracted from English Books (800M words) and Wikipedia pages (2,500M words). The pretrained BERT model is able to represent input sentences in a way that captures the underlying syntax, semantics, meanings and relationships between the words. The pretrained BERT model has been open-sourced and made publicly available. It can be easily reused for other tasks such as classification. Figure~\ref{BERT_class} shows how BERT can be used for sentiment classification. The output at the first position (corresponding to the [CLS] token) can be used as the input of a classifier to determine whether the input textual description is positive or negative. If we have a multiclass classification problem, we can tweak the classifier so that it has more output neurons.

\begin{figure*}[]
\centering
\begin{center}
\includegraphics[scale=0.3]{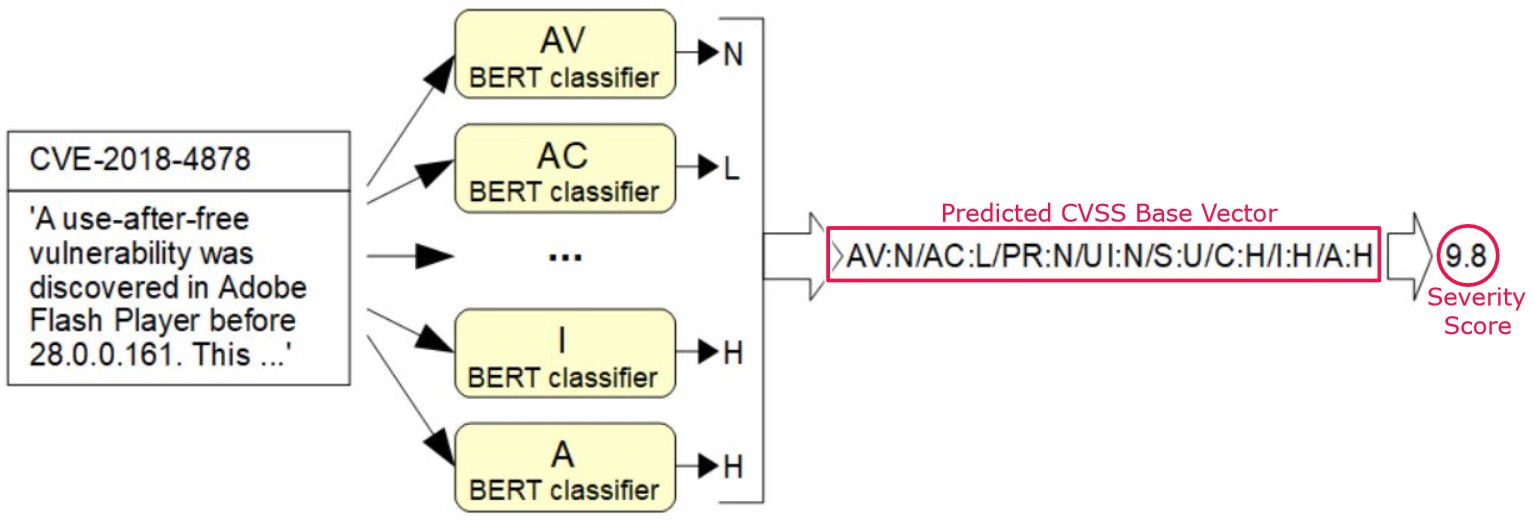}
\end{center}
\caption{CVSS vector and severity score prediction pipeline}
\label{CVSS_predictor_pipeline}
\end{figure*}

\section{Common Vulnerability Scoring System} \label{CVSS}

Common Vulnerability Scoring System (CVSS)~\cite{CVSSv3.1} is a standard to describe the principal characteristics of  a vulnerability and assess its severity relative to other vulnerabilities. Multiple versions of the standard have been released CVSS v2, v3.0 and v3.1. For our work, we use CVSS v3.1, the latest version of the standard at the time of writing. For the rest of this paper CVSS refers to CVSS v3.1, unless specified otherwise. CVSS captures and summarizes the characteristics of a vulnerability in a vector composed of three metric groups: Base, Temporal, and Environmental.

The Base metrics reflects the severity of a vulnerability according to its intrinsic characteristics which are constant over time and across different environments. The Temporal metrics adjust the Base severity of a vulnerability based on factors that change over time, such as the availability of exploit code. The Environmental metrics adjust the Base and Temporal metrics to a specific computing environment (taking into account factors such as the presence of mitigations in that environment). In fact, temporal and environmental metrics are rarely used in practice. Only the CVSS Base metrics are adopted by NVD to provide severity analysis of vulnerabilities. For the rest of this paper, when we refer to CVSS vectors or metrics, we specifically refer to CVSS Base vectors and metrics. 

The CVSS Base vector consists of two sets of metrics: the Exploitability metrics and the Impact metrics. Exploitability metrics characterize the ease and technical means by which a vulnerability can be exploited. Impact metrics reflect the consequences of a successful exploit of the vulnerability on the impacted component. Table~\ref{tab:cvss_metrics} describes the different Exploitability and Impact metrics that compose the CVSS Base vector. It also provides the value that each of these metric can take.

A vulnerability is publicly disclosed through the CVE system. It is identified by an ID and added to the CVE list. For each new CVE entry, NVD analysts assign values to the different metrics of the CVSS Base vector. The assigned values then goes through different equations to calculate a severity score ranging from 0.0 to 10.0. The details of how to compute severity scores from Base vectors is described in the CVSS specification document~\cite{CVSSv3.1}. For example, the CVSS vector assigned to CVE-2018-4878, a use-after-free vulnerability discovered in Adobe Flash Player~\cite{CVE-2018-4878}, is:
\begin{center}
    AV:N/AC:L/PR:N/UI:N/S:U/C:H/I:H/A:H
\end{center}
Where AV:N stands for Attack Vector metric is equal to Network, AC:L stands for Attack Complexity metric is equal to Low, and so on (See Table~\ref{tab:cvss_metrics} for further detail about the different metrics). The severity score corresponding to this vector is 9.8 (critical).

\section{CVSS Vector and Severity Score Prediction}

\subsection{Model Description}

To predict the Base CVSS vector from the description of a vulnerability, we propose to train multiple BERT classifiers, one for each metric composing the CVSS vector (AV, AC, PR, etc). As illustrated in Figure~\ref{CVSS_predictor_pipeline}, the textual description of a vulnerability (provided by the CVE system) is passed to each trained BERT classifier to determine the value of the different metrics that compose the CVSS vector. The individual values predicted by each classifier are then concatenated to get the full predicted CVSS vector. From the predicted CVSS vector we can infer the severity score (using the equations provided by the CVSS specification document).

\subsection{Experimental Setup}

We use data provided by the NVD database~\cite{NVD_NIST}. NVD database is fully synchronized with the MITRE CVE list and each vulnerability is identified by a CVE ID. For each CVE vulnerability the database includes a textual description (that we use as input of our BERT classifiers), a CVSS vector (that we use as outputs to train our BERT classifiers), and a severity score. Our full dataset consists of 3 years of CVE vulnerability data, from 2018 to 2020, corresponding to a total of 45,926 samples. The full dataset is split randomly into a training set and a test set, containing 22,963 CVE vulnerabilities each.

We use BERT-small, a lighter version of BERT proposed in \cite{turc2019well}, consisting of 4 transformer encoder layers (instead of 12 for the original BERT Base model) and a hidden embedding size of 512 (instead of 768 for BERT Base). With fewer parameters than the original BERT Base model, BERT-small is less computationally expensive and faster to train. The vulnerability descriptions are tokenized using the pretrained BERT-small tokenizer. Padding and truncation are used so as to have token sequences of  length 128. 

We train multiple BERT models for classification tasks, one for each individual metric of the CVSS Base vector.  As explained in Section \ref{BERT}, BERT is a pretrained model. That is, it was trained on huge corpus of textual data to effectively model texts written in English (underlying syntax, semantics, meanings, relationships between words, etc.). To have a BERT classifier, we add a classifier on top of the pretrained BERT model. We have to be careful during training because the added classifier is randomly initialized. Hence, very large weight updates will be propagated through the network, and the representation learned by the pretrained BERT model will be destroyed. To avoid this issue, the weights of the pretrained BERT model are frozen for the first 3 epochs and only the weights of the classifier are fine-tuned. After the weights of the classifier reach reasonable values, the weights of the pretrained BERT model are unfrozen. The classifier and the BERT model are jointly trained for another 3 epochs.

\subsection{Experimental Results}

\begin{table}[]
\centering
\caption{Performance of the BERT classifiers on the test set}
\label{tab:BERT_performance}
\begin{tabular}{|c|c|c|c|c|}
\hline
            & \textbf{Accuracy} & \textbf{\begin{tabular}[c]{@{}c@{}}Precision\\  (weighted)\end{tabular}} & \textbf{\begin{tabular}[c]{@{}c@{}}Recall\\  (weighted)\end{tabular}} & \textbf{\begin{tabular}[c]{@{}c@{}}F1-score\\ (weighted)\end{tabular}} \\ \hline
\textbf{AV} & 0.9115            & 0.9090                                                                   & 0.9115                                                                & 0.9089                                                                 \\ \hline
\textbf{AC} & 0.9607            & 0.9570                                                                   & 0.9607                                                                & 0.9574                                                                 \\ \hline
\textbf{PR} & 0.8379            & 0.8392                                                                   & 0.8379                                                                & 0.8378                                                                 \\ \hline
\textbf{UI} & 0.9321            & 0.9318                                                                   & 0.9321                                                                & 0.9319                                                                 \\ \hline
\textbf{S}  & 0.9545            & 0.9553                                                                   & 0.9545                                                                & 0.9548                                                                 \\ \hline
\textbf{C}  & 0.8704            & 0.8714                                                                   & 0.8704                                                                & 0.8681                                                                 \\ \hline
\textbf{I}  & 0.8735            & 0.8736                                                                   & 0.8735                                                                & 0.8731                                                                 \\ \hline
\textbf{A}  & 0.8894            & 0.8868                                                                   & 0.8894                                                                & 0.8863                                                                 \\ \hline
\end{tabular}
\end{table}

Table~\ref{tab:BERT_performance} presents the performance of the different BERT classifiers on the test set. In order to take class imbalance into account, weighted average is used to compute precision, recall and F1-score. All classifiers achieve relatively high accuracy, ranging from 83.79\% to 96.07\%. The easiest CVSS metrics to predict are Attack Complexity (AC) and Scope (S), with an achieved accuracy on the test set of 96.07\% and 95.45\% respectively. Note however that the high performance for AC metric can be partly explained by the highly imbalanced classes for this specific metric, with 93\% of the samples belonging to one class. The CVSS metric most difficult to predict is Privileges Required (PR) with an accuracy of 83.79\%.

From the predicted CVSS Base vectors, we compute the CVSS severity scores for each CVE vulnerability in the test set. The Mean Squared Error ($MSE$) and Mean Absolute Error ($MAE$) between the predicted severity scores and the true severity scores is of 1.79 and 0.73 respectively. The predicted severity scores exactly match the true severity scores ($MAE=0$) for 55.3\% of the CVE vulnerabilities in the test set. The MAE is also less than 1 for 75\% of the vulnerabilities in the test set.

\section{Explainability}

\textit{Interpretability} of a machine learning model refers to the ability to determine the cause and effect relationship between the inputs of a model and its prediction. It allows a human user to understand and explain the decision of a model~\cite{molnar2020interpretable}. It also allows knowledge discovery (spot specific patterns in the data) and helps debug the model, for example, to better understand incorrect predictions. Depending on the context, \textit{interpretability} and \textit{explainability} might refers to slightly different concepts. In this paper, we use both terms interchangeably.

\begin{table}[]
\ttfamily
\caption{Examples of predictions along with the top 5 most relevant tokens (as determined by gradient based input saliency) in bold and underlined}
\label{tab:interpretability_examples}
\resizebox{\columnwidth}{!}{
\begin{tabular}{|l|}
\hline
\begin{tabular}[c]{@{}l@{}}CVE-2020-9804\\ Predicted Attack Vector (AV): PHYSICAL (P)\\ \\ A logic issue was addressed with improved restrictions.\\ This issue is fixed in macOS Cat\underline{\textbf{a}}lina 10.15.5. Insert\underline{\textbf{ing}}\\ a \underline{\textbf{USB device}} that sends invalid \underline{\textbf{messages}} may cause a\\ kernel panic.\end{tabular} \\ \hline
\begin{tabular}[c]{@{}l@{}}CVE-2018-15611\\ Predicted Privileges Required (PR): HIGH (H)\\ \\ A vulnerability in the local system administration\\ component of Avaya Aura \underline{\textbf{Communication}} Manager can\\ allow an authentic\underline{\textbf{ated, privileged user}} on the local\\ system to gain root privileges. Affected versions include\\ 6.3.x and all 7.x version prior to 7.1.3.1.\end{tabular} \\ \hline
\begin{tabular}[c]{@{}l@{}}CVE-2019-12773\\ Predicted User Interaction (UI): REQUIRED (R)\\ \\ An issue was discovered in Verint Impact 360 15.1. At\\ wfo/help/help\_popup.jsp, the help URL parameter can\\ be changed to embed arbitrary content inside of an\\ iFrame. Attackers may use this in conjunction with\\ social engineering to embed malicious scripts or\\ phishing pages on a site where this product is\\ installed, given the attacker can \underline{\textbf{convince}} a \underline{\textbf{victim}} to\\ \underline{\textbf{visit}} a \underline{\textbf{crafted link}}.\end{tabular} \\ \hline
\begin{tabular}[c]{@{}l@{}}CVE-2019-16278\\ Predicted Confidentiality Impact (C): HIGH (H)\\ \\ \underline{\textbf{IBM Financial}} Transaction Manager for SWIFT Services\\ for Multiplatforms 3.2.4 could allow an remote attacker\\ to obtain sensitive information, caused by a man in the \\ \underline{\textbf{middle}} attack. By SSL striping, an attacker could \\ exploit this vulnerability to obtain  \underline{\textbf{sensitive}} \\ \underline{\textbf{information}}.\end{tabular} \\ \hline
\begin{tabular}[c]{@{}l@{}}CVE-2019-9964\\ Predicted Availability Impact (A): HIGH (H)\\ \\ XnView MP 0.93.1 on \underline{\textbf{Windows}} allows remote \\ attackers to \underline{\textbf{cause}} a \underline{\textbf{denial}} of \underline{\textbf{service}} (application \underline{\textbf{crash}})\\ or possibly have unspecified other impact via a crafted\\ file, related to ntdll!RtlpNtMakeTemporaryKey.\end{tabular} \\ \hline
\end{tabular}}
\end{table}

\subsection{Gradient-based input saliency}

Gradient-based input saliency methods can be used to find out which input tokens (sub-words) are the most important for a given prediction made by the model~\cite{alammar2020explaining,bastings2020elephant}. Note that it is also possible to determine the importance of each token based on the attention weights~\cite{wiegreffe2019attention}. The assumption is that input tokens accorded high attention are responsible for the model output. However recent studies cast doubts to the degree to which attention weights provide meaningful explanations for predictions~\cite{bastings2020elephant, jain2019attention}. In~\cite{bastings2020elephant},  J. Bastings et al. argue that saliency methods are better suited to determine what inputs are the most relevant to predictions. In~\cite{atanasova2020diagnostic}, P. Atanasova et al. compare different explainability techniques and show that gradient-based input saliency methods perform the best.

To predict each metric of the CVSS vector, we trained different BERT models for classification task. That is, the output layer of our models consist of a set of logits, each corresponding to a specific class, and a softmax function is used to transform those logits into probabilities that sum up to one. The class with the highest probability is predicted by the model.

For a given prediction, we can determine how important each input token is to the prediction by calculating the gradient of the score (value of the output logit) corresponding to the predicted class, with respect to the inputs. Specifically, the smallest change in the input token with the highest gradient-based saliency value will result in a large change in the output of the model. 
We use the \textit{Gradient X Input} method in which the computed gradient vector per token is multiplied by the input embedding of the token. Taking the $L2$ norm of the resulting vector gives the token's feature importance score, a measure of how sensitive the model is to that specific input token. More formally, the importance of the token at the $i^{th}$ position in the input sequence is given by:
\begin{center}
    $\left \| \nabla_{X_{i}} f_{c}(X_{1:n}) \cdot X_{i} \right \|_{2}$
\end{center}
where:
\begin{itemize}
    \item $X_{i}$ is the embedding vector of the $i^{th}$ input token
    \item $X_{1:n}$ is the list of embedding vectors of all the tokens in the input sequence (of length $n$)
    \item $f_{c}(X_{1:n})$ is the score of the predicted class after a forward pass through the model.
    \item $\nabla_{X_{i}} f_{c}(X_{1:n})$ is the back-propagated gradient of the score of the predicted class. 
\end{itemize}

\begin{table*}[]
\caption{Tokens most often associated to specific values of the CVSS metrics}
\label{tab:most_relevant_words}
\resizebox{\textwidth}{!}{
\begin{tabular}{cc|l|l|}
\cline{3-4}
                                                   &            & \multicolumn{1}{c|}{\textbf{Unigrams}}                                                                                                                                         & \multicolumn{1}{c|}{\textbf{Bigrams}}                                                                                                                        \\ \hline
\multicolumn{1}{|c|}{\multirow{4}{*}{\textbf{AV}}} & \textbf{N} & 'remote', 'xss', 'php', 'web', 'network', 'http', 'script', 'file', 'ur', 'site'                                                                                               & 'network access', 'site script', 'with network', 'google chrome', 'remote attacker'                                                                          \\ \cline{2-4} 
\multicolumn{1}{|c|}{}                             & \textbf{A} & \begin{tabular}[c]{@{}l@{}}'adjacent', 'attacker', 'cisco', 'network', 'via', 'route', 'same', 'access', 'intel',\\ 'allows'\end{tabular}                                      & \begin{tabular}[c]{@{}l@{}}'adjacent attacker', 'via adjacent', 'same network', 'adjacent access',\\ 'adjacent attackers'\end{tabular}                       \\ \cline{2-4} 
\multicolumn{1}{|c|}{}                             & \textbf{L} & \begin{tabular}[c]{@{}l@{}}'local', 'privilege', 'windows', 'privileges', 'file', 'kernel', 'elevation', 'remote',\\ 'access', 'attacker\end{tabular}                          & \begin{tabular}[c]{@{}l@{}}'local attacker', 'local access', 'local users', 'local information',\\ 'infrastructure where'\end{tabular}                       \\ \cline{2-4} 
\multicolumn{1}{|c|}{}                             & \textbf{P} & \begin{tabular}[c]{@{}l@{}}'physical', 'access', 'usb', 'device', 'allows', 'physically', 'intel', 'attacker', 'via',\\ 'lock'\end{tabular}                                    & \begin{tabular}[c]{@{}l@{}}'physical access', 'via physical', 'usb device', 'allows physical',\\ 'allows physically','malicious usb'\end{tabular}            \\ \hline
\multicolumn{1}{|c|}{\multirow{2}{*}{\textbf{AC}}} & \textbf{L} & 'xss', 'file', 'crafted', 'web', 'user', 'site', 'needed', 'server', 'easily', 'exploit'                                                                                       & 'easily exploit', '. easily', 'user interaction', 'site script', 'successful exploitation'                                                                   \\ \cline{2-4} 
\multicolumn{1}{|c|}{}                             & \textbf{H} & \begin{tabular}[c]{@{}l@{}}'engine', 'difficult', 'exploit', 'vulnerability', 'script', 'to', 'race', 'middle', 'man',\\ 'memory'\end{tabular}                                 & 'difficult to', 'to exploit', 'race condition', '. difficult', 'exploit vulnerability'                                                                       \\ \hline
\multicolumn{1}{|c|}{\multirow{3}{*}{\textbf{PR}}} & \textbf{H} & \begin{tabular}[c]{@{}l@{}}'privileged', 'high', 'allows', 'attacker', 'admin', 'user', 'authentic', 'administrator',\\ 'privileges', 'oracle'\end{tabular}                    & \begin{tabular}[c]{@{}l@{}}'high privileged', 'privileged attacker', 'allows high', 'privileged user',\\ 'system execution'\end{tabular}                     \\ \cline{2-4} 
\multicolumn{1}{|c|}{}                             & \textbf{L} & \begin{tabular}[c]{@{}l@{}}'authenticated', 'local', 'user', 'users', 'allows', 'attacker', 'low', 'elevation',\\ 'privileged', 'privilege'\end{tabular}                       & 'low privileged', 'allows low', 'local users', 'authenticated user', 'allows local'                                                                          \\ \cline{2-4} 
\multicolumn{1}{|c|}{}                             & \textbf{N} & \begin{tabular}[c]{@{}l@{}}'unauthenticated', 'remote', 'attackers', 'attacker', 'exploitation', 'code',\\ 'corruption', 'vulnerability', 'network', 'successful'\end{tabular} & \begin{tabular}[c]{@{}l@{}}'remote attackers', 'successful exploitation', 'unauthenticated attacker',\\ 'remote code', 'remote attacker'\end{tabular}        \\ \hline
\multicolumn{1}{|c|}{\multirow{2}{*}{\textbf{UI}}} & \textbf{R} & 'xss', 'site', 'script', 'cross', 'crafted', 'malicious', 'interaction', 'file', 'csrf', 'php'                                                                                 & 'site script', 'human interaction', 'crafted html', 'html page', 'xss via'                                                                                   \\ \cline{2-4} 
\multicolumn{1}{|c|}{}                             & \textbf{N} & \begin{tabular}[c]{@{}l@{}}'local', 'network', 'user', 'server', 'oracle', 'sql', 'allows', 'devices', 'php', \\ 'injection'\end{tabular}                                      & \begin{tabular}[c]{@{}l@{}}'sql injection', 'not needed', 'contract implementation', 'linux kernel',\\ 'network access'\end{tabular}                         \\ \hline
\multicolumn{1}{|c|}{\multirow{2}{*}{\textbf{S}}}  & \textbf{C} & 'xss', 'script', 'site', 'cross', 'impact', 'products', 'attacks', 'stored', 'may', 'has'                                                                                      & 'site script', 'xss via', 'attacks may', 'stored xss', 'has xss'                                                                                             \\ \cline{2-4} 
\multicolumn{1}{|c|}{}                             & \textbf{U} & \begin{tabular}[c]{@{}l@{}}'code', 'memory', 'execution', 'information', 'buffer', 'function', 'can', 'denial',\\ 'sql', 'password'\end{tabular}                               & \begin{tabular}[c]{@{}l@{}}'code execution', 'can result', 'information disclosure', 'arbitrary code',\\ 'sql injection'\end{tabular}                        \\ \hline
\multicolumn{1}{|c|}{\multirow{3}{*}{\textbf{C}}}  & \textbf{H} & \begin{tabular}[c]{@{}l@{}}'code', 'execution', 'arbitrary', 'execute', 'disclosure', 'privilege', 'privileges',\\ 'injection', 'information', 'bounds'\end{tabular}           & \begin{tabular}[c]{@{}l@{}}'code execution', 'arbitrary code', 'execute arbitrary', 'remote code',\\ 'information disclosure'\end{tabular}                   \\ \cline{2-4} 
\multicolumn{1}{|c|}{}                             & \textbf{L} & \begin{tabular}[c]{@{}l@{}}'xss', 'script', 'arbitrary', 'html', 'site', 'stored', 'unauthorized', 'allows', 'web', \\ 'code'\end{tabular}                                     & 'site script', 'stored xss', 'unauthorized read', 'allows xss', 'read access'                                                                                \\ \cline{2-4} 
\multicolumn{1}{|c|}{}                             & \textbf{N} & \begin{tabular}[c]{@{}l@{}}'denial', 'service', 'contract', 'cause', 'dos', 'crash', 'result', 'balance', 'set', \\ 'function'\end{tabular}                                    & \begin{tabular}[c]{@{}l@{}}'frequently repeat', 'complete dos', 'contract implementation', 'null pointer',\\ 'service attack'\end{tabular}                   \\ \hline
\multicolumn{1}{|c|}{\multirow{3}{*}{\textbf{I}}}  & \textbf{H} & \begin{tabular}[c]{@{}l@{}}'code', 'execution', 'arbitrary', 'execute', 'privilege', 'injection', 'privileges',\\ 'remote', 'vulnerability', 'sql'\end{tabular}                & \begin{tabular}[c]{@{}l@{}}'code execution', 'arbitrary code', 'execute arbitrary', 'remote code',\\ 'sql injection'\end{tabular}                            \\ \cline{2-4} 
\multicolumn{1}{|c|}{}                             & \textbf{L} & 'xss', 'script', 'site', 'html', 'insert', 'update', 'unauthorized', 'php', 'stored', 'result'                                                                                 & 'site script', 'unauthorized update', 'stored xss', 'update ,', ', insert'                                                                                   \\ \cline{2-4} 
\multicolumn{1}{|c|}{}                             & \textbf{N} & \begin{tabular}[c]{@{}l@{}}'denial', 'disclosure', 'information', 'read', 'service', 'crash', 'vulnerability', 'dos',\\ 'sensitive', 'and'\end{tabular}                        & \begin{tabular}[c]{@{}l@{}}'information disclosure', 'sensitive information', 'read vulnerability',\\ 'disclosure vulnerability', 'bounds read'\end{tabular} \\ \hline
\multicolumn{1}{|c|}{\multirow{3}{*}{\textbf{A}}}  & \textbf{H} & \begin{tabular}[c]{@{}l@{}}'code', 'service', 'execution', 'denial', 'execute', 'arbitrary', 'buffer', 'crash', 'of',\\ 'privilege'\end{tabular}                               & \begin{tabular}[c]{@{}l@{}}'code execution', 'arbitrary code', 'denial of', 'of service',\\ 'execute arbitrary', 'buffer over'\end{tabular}                  \\ \cline{2-4} 
\multicolumn{1}{|c|}{}                             & \textbf{L} & \begin{tabular}[c]{@{}l@{}}'partial', 'cause', 'denial', 'service', 'dos', 'consume', 'entity', 'processing', 'restart',\\ 'injection'\end{tabular}                            & \begin{tabular}[c]{@{}l@{}}'partial denial', 'entity injection', 'undefined behavior', 'temporarily unavailable',\\ 'or consume'\end{tabular}                \\ \cline{2-4} 
\multicolumn{1}{|c|}{}                             & \textbf{N} & \begin{tabular}[c]{@{}l@{}}'xss', 'disclosure', 'information', 'script', 'site', 'read', 'data', 'accessible', 'files',\\ 'arbitrary'\end{tabular}                             & \begin{tabular}[c]{@{}l@{}}'site script', 'information disclosure', 'accessible data', 'stored xss',\\ 'sensitive information'\end{tabular}                  \\ \hline
\end{tabular}}
\end{table*}

%\underline{\textbf{}}
Table~\ref{tab:interpretability_examples} presents examples of predictions made by our trained BERT classifiers on samples from the test set, along with the top 5 most important tokens (as determined by gradient based input saliency) for each prediction. Note that tokens can be sub-words and also include punctuation marks. It is interesting to see that the top 5 most relevant tokens for each prediction include terms in agreement with the rationales of a human cybersecurity expert. For example, to determine that the Attack Vector (AV) for CVE-2020-9804 is Physical (P), the most important terms for a human expert are 'USB device'. Those terms are also the top most relevant tokens as determined by gradient based input saliency. Similarly, to predict that User Interaction (UI) is Required (R) for CVE-2019-12773, the top most relevant tokens include terms like 'convince', 'visit', 'crafted link'. Those terms are in line with what a human expert would consider important to correctly classify the vulnerability. Because top most relevant tokens as determined by gradient based input saliency often include terms that are also important for a human expert, it is easy for an end-user to make sense of it.

\subsection{Tokens Most Often Associated with Specific Values of the CVSS Metrics}

Table~\ref{tab:most_relevant_words} lists for the different values of each CVSS metric, the most relevant input tokens. That is, the input tokens that are most often associated by the classifiers to a specific prediction. The methodology to determine those tokens for each classifier is as follows:
\begin{itemize}
    \item Keep only samples in the test set for which the classifier predicts the class with high confidence (the output after softmax for the predicted class is greater than 0.9).
    \item For each vulnerability description determine the top 5 input tokens in terms of their importance to the predicted class using the gradient-based input saliency method.
    \item For each possible class, create a list of the most important tokens by concatenating the top 5 tokens of all vulnerabilities belonging to that class.
    \item Compute the number of occurrences of each token in that list to determine the top 10 tokens that are most often associated by the classifier to that specific class.
\end{itemize}
Some tokens corresponding to sub-words were completed to represent valid meaningful words. For example, the word 'xss' (shorthand for \textit{cross-site scripting}, a type of vulnerability usually found in web applications) is represented by two tokens 'x' and 'ss'.

Table~\ref{tab:most_relevant_words} also shows bigrams most often associated by a classifier to a specific class. A bigram is a pair of consecutive tokens. A relevant bigram occurs in a vulnerability description if two consecutive tokens are both among the top 5 most important tokens. For example, if in a description both 'network' and 'access' are among the top 5 input tokens in terms of their importance to the prediction (as determined by gradient-based input saliency) and they also form two consecutive tokens, then 'network access' is counted as a bigram relevant to the prediction. As with individual input tokens (unigrams), we count the number of occurrences of each relevant bigram over all the vulnerability descriptions and all classes in the test set to determine the top 5 bigrams most often associated by a classifier with a specific class.

The tokens most often associated with the different values of the CVSS metrics are in agreement with the reasoning of a human cybersecurity expert. For example, for the Attack Vector (AV) metric, the terms 'adjacent attacker' are correctly associated with the class Adjacent (A), while the terms 'physical' and 'usb' are associated with the class Physical (P). For Attack Complexity (AC) metric, the expression 'easily exploit' is associated with Low (L), while 'difficult to' is associated with High (H). Similarly, for the Privileges Required (PR) metric, the expressions 'high privileged' and 'low privileges' are precisely associated with High (H) and Low (L) respectively. For the User Interaction (UI) metric, the terms 'xss', 'site script' (vulnerability that requires the victim to visit a compromised website) and 'human interaction' are accurately associated with Required (R). The expression 'information disclosure' is related to a Confidentiality impact (C) metric equal to High (H). For the Availability impact (A) metric, the terms 'denial' and 'crash' are correctly associated with High (H).
Often, the tokens most often associated with a particular value of a CVSS metric include terms that are in line with the rationales of a human cybesecurity expert, making the explanations provided by gradient-based input saliency method comprehensible for end-users.

\section{Conclusion}

Each year, thousands of computer security vulnerabilities are disclosed. To address the security threat posed by those vulnerabilities organisations must prioritize their efforts and allocate their limited resources effectively. Knowing the severity of a vulnerability might help them to determine which vulnerabilities should be addressed first. However, when a new vulnerability is disclosed, only a textual description of it is available. Cybersecurity experts later provide an analysis of the severity of the vulnerability using the CVSS standard. The characteristics of a vulnerability are summarized into a CVSS vector from which a severity score can be computed. The process of attributing a CVSS vector and severity score to a vulnerability requires lot of time and manpower.

We proposed to leverage recent advances in NLP to automatically determine the CVSS vector and severity score of a vulnerability from its description. To this purpose we trained multiple BERT classifiers, one for each metric composing the CVSS vector. Experimental results show that the classifiers achieve high accuracy. The values predicted by each individual classifier are concatenated to construct the CVSS vector, from which a numerical severity score is computed. The predicted severity score is very close to the real severity score provided by human experts. For explainability purpose, gradient-based input saliency method was used to determine the most relevant input words for a given prediction made by our classifiers. The top relevant words often include terms in agreement with the rationales of a human cybersecurity expert, making the explanation comprehensible for end-users.

\bibliographystyle{IEEEtran}

\bibliography{bib}

\begin{thebibliography}{10}
\providecommand{\url}[1]{#1}
\csname url@samestyle\endcsname
\providecommand{\newblock}{\relax}
\providecommand{\bibinfo}[2]{#2}
\providecommand{\BIBentrySTDinterwordspacing}{\spaceskip=0pt\relax}
\providecommand{\BIBentryALTinterwordstretchfactor}{4}
\providecommand{\BIBentryALTinterwordspacing}{\spaceskip=\fontdimen2\font plus
\BIBentryALTinterwordstretchfactor\fontdimen3\font minus
  \fontdimen4\font\relax}
\providecommand{\BIBforeignlanguage}[2]{{%
\expandafter\ifx\csname l@#1\endcsname\relax
\typeout{** WARNING: IEEEtran.bst: No hyphenation pattern has been}%
\typeout{** loaded for the language `#1'. Using the pattern for}%
\typeout{** the default language instead.}%
\else
\language=\csname l@#1\endcsname
\fi
#2}}
\providecommand{\BIBdecl}{\relax}
\BIBdecl

\bibitem{CVE_MITRE}
\BIBentryALTinterwordspacing
\emph{{Common Vulnerabilities and Exposures (CVE)}}, {(accessed July 27,
  2021)}. [Online]. Available: \url{https://cve.mitre.org/}
\BIBentrySTDinterwordspacing

\bibitem{NVD_NIST}
\BIBentryALTinterwordspacing
\emph{{National Vulnerability Database (NVD)}}, {(accessed July 27, 2021)}.
  [Online]. Available: \url{https://nvd.nist.gov/}
\BIBentrySTDinterwordspacing

\bibitem{CVSSv3.1}
\BIBentryALTinterwordspacing
\emph{{Common Vulnerability Scoring System version 3.1: Specification
  Document}}, {2019 (accessed July 27, 2021)}. [Online]. Available:
  \url{https://www.first.org/cvss/specification-document}
\BIBentrySTDinterwordspacing

\bibitem{devlin2018bert}
J.~Devlin, M.-W. Chang, K.~Lee, and K.~Toutanova, ``Bert: Pre-training of deep
  bidirectional transformers for language understanding,'' \emph{arXiv preprint
  arXiv:1810.04805}, 2018.

\bibitem{elbaz2020fighting}
C.~Elbaz, L.~Rilling, and C.~Morin, ``Fighting n-day vulnerabilities with
  automated cvss vector prediction at disclosure,'' in \emph{Proceedings of the
  15th International Conference on Availability, Reliability and Security},
  2020, pp. 1--10.

\bibitem{khazaei2016automatic}
A.~Khazaei, M.~Ghasemzadeh, and V.~Derhami, ``An automatic method for cvss
  score prediction using vulnerabilities description,'' \emph{Journal of
  Intelligent \& Fuzzy Systems}, vol.~30, no.~1, pp. 89--96, 2016.

\bibitem{han2017learning}
Z.~Han, X.~Li, Z.~Xing, H.~Liu, and Z.~Feng, ``Learning to predict severity of
  software vulnerability using only vulnerability description,'' in \emph{2017
  IEEE International conference on software maintenance and evolution
  (ICSME)}.\hskip 1em plus 0.5em minus 0.4em\relax IEEE, 2017, pp. 125--136.

\bibitem{zong2019analyzing}
S.~Zong, A.~Ritter, G.~Mueller, and E.~Wright, ``Analyzing the perceived
  severity of cybersecurity threats reported on social media,'' \emph{arXiv
  preprint arXiv:1902.10680}, 2019.

\bibitem{zou2019autocvss}
D.~Zou, J.~Yang, Z.~Li, H.~Jin, and X.~Ma, ``Autocvss: An approach for
  automatic assessment of vulnerability severity based on attack process,'' in
  \emph{International Conference on Green, Pervasive, and Cloud
  Computing}.\hskip 1em plus 0.5em minus 0.4em\relax Springer, 2019, pp.
  238--253.

\bibitem{tavabi2018darkembed}
N.~Tavabi, P.~Goyal, M.~Almukaynizi, P.~Shakarian, and K.~Lerman, ``Darkembed:
  Exploit prediction with neural language models,'' in \emph{Proceedings of the
  AAAI Conference on Artificial Intelligence}, vol.~32, no.~1, 2018.

\bibitem{russo2019summarizing}
E.~R. Russo, A.~Di~Sorbo, C.~A. Visaggio, and G.~Canfora, ``Summarizing
  vulnerabilities’ descriptions to support experts during vulnerability
  assessment activities,'' \emph{Journal of Systems and Software}, vol. 156,
  pp. 84--99, 2019.

\bibitem{jacobs2019exploit}
J.~Jacobs, S.~Romanosky, B.~Edwards, M.~Roytman, and I.~Adjerid, ``Exploit
  prediction scoring system (epss),'' \emph{arXiv preprint arXiv:1908.04856},
  2019.

\bibitem{vaswani2017attention}
A.~Vaswani, N.~Shazeer, N.~Parmar, J.~Uszkoreit, L.~Jones, A.~N. Gomez,
  {\L}.~Kaiser, and I.~Polosukhin, ``Attention is all you need,'' in
  \emph{Advances in neural information processing systems}, 2017, pp.
  5998--6008.

\bibitem{CVE-2018-4878}
\BIBentryALTinterwordspacing
\emph{{CVE-2018-4878 Detail}}, {(accessed July 27, 2021)}. [Online]. Available:
  \url{https://nvd.nist.gov/vuln/detail/cve-2018-4878}
\BIBentrySTDinterwordspacing

\bibitem{turc2019well}
I.~Turc, M.-W. Chang, K.~Lee, and K.~Toutanova, ``Well-read students learn
  better: On the importance of pre-training compact models,'' \emph{arXiv
  preprint arXiv:1908.08962}, 2019.

\bibitem{molnar2020interpretable}
C.~Molnar, \emph{Interpretable machine learning}.\hskip 1em plus 0.5em minus
  0.4em\relax Lulu. com, 2020.

\bibitem{alammar2020explaining}
\BIBentryALTinterwordspacing
J.~Alammar, ``Interfaces for explaining transformer language models,'' 2020.
  [Online]. Available:
  \url{https://jalammar.github.io/explaining-transformers/}
\BIBentrySTDinterwordspacing

\bibitem{bastings2020elephant}
J.~Bastings and K.~Filippova, ``The elephant in the interpretability room: Why
  use attention as explanation when we have saliency methods?'' \emph{arXiv
  preprint arXiv:2010.05607}, 2020.

\bibitem{wiegreffe2019attention}
S.~Wiegreffe and Y.~Pinter, ``Attention is not not explanation,'' in
  \emph{Proceedings of the 2019 Conference on Empirical Methods in Natural
  Language Processing and the 9th International Joint Conference on Natural
  Language Processing (EMNLP-IJCNLP)}, 2019, pp. 11--20.

\bibitem{jain2019attention}
S.~Jain and B.~C. Wallace, ``Attention is not explanation,'' in
  \emph{Proceedings of the 2019 Conference of the North American Chapter of the
  Association for Computational Linguistics: Human Language Technologies,
  Volume 1 (Long and Short Papers)}, 2019, pp. 3543--3556.

\bibitem{atanasova2020diagnostic}
P.~Atanasova, J.~G. Simonsen, C.~Lioma, and I.~Augenstein, ``A diagnostic study
  of explainability techniques for text classification,'' in \emph{Proceedings
  of the 2020 Conference on Empirical Methods in Natural Language Processing
  (EMNLP)}, 2020, pp. 3256--3274.

\end{thebibliography}

\end{document}